\documentclass[10pt,twocolumn,letterpaper]{article}

\usepackage{cvpr}              
\usepackage[accsupp]{axessibility}
\usepackage{amssymb}
\usepackage{pifont}
\newcommand{\cmark}{\textrm{\ding{51}}}%
\newcommand{\xmark}{\textrm{\ding{55}}}%
\newcommand{\naline}{\raisebox{0.5ex}{\rule{0.5cm}{1pt}}}
\usepackage[dvipsnames]{xcolor}

\definecolor{cvprblue}{rgb}{0.21,0.49,0.74}
\usepackage[pagebackref,breaklinks,colorlinks,citecolor=cvprblue]{hyperref}

\def\paperID{*****} 
\def\confName{CVPR}
\def\confYear{2024}

\title{AutoSoccerPose: Automated 3D posture Analysis of Soccer Shot Movements}

\author{
Calvin Yeung \text{  } \text{  } \text{  }\text{  }\text{  }\text{  }\text{  } Kenjiro Ide\\
Nagoya University, Japan\\
{\tt\small yeung.chikwong@g.sp.m.is.nagoya-u.ac.jp}\\
{\tt\small ide.kenjiro@g.sp.m.is.nagoya-u.ac.jp}
\and
Keisuke Fujii\\
Nagoya University, Japan\\
RIKEN/JST PRESTO, Japan\\
{\tt\small fujii@i.nagoya-u.ac.jp}
}

\begin{document}
\maketitle
\begin{abstract}
Image understanding is a foundational task in computer vision, with recent applications emerging in soccer posture analysis. However, existing publicly available datasets lack comprehensive information, notably in the form of posture sequences and 2D pose annotations. Moreover, current analysis models often rely on interpretable linear models (e.g., PCA and regression), limiting their capacity to capture non-linear spatiotemporal relationships in complex and diverse scenarios. To address these gaps, we introduce the 3D Shot Posture (3DSP) dataset in soccer broadcast videos, which represents the most extensive sports image dataset with 2D pose annotations to our knowledge. Additionally, we present the 3DSP-GRAE (Graph Recurrent AutoEncoder) model, a non-linear approach for embedding pose sequences. Furthermore, we propose AutoSoccerPose, a pipeline aimed at semi-automating 2D and 3D pose estimation and posture analysis. While achieving full automation proved challenging, we provide a foundational baseline, extending its utility beyond the scope of annotated data. We validate AutoSoccerPose on SoccerNet and 3DSP datasets, and present posture analysis results based on 3DSP. The dataset, code, and models are available at: \url{https://github.com/calvinyeungck/3D-Shot-Posture-Dataset}.

\end{abstract}


\section{Introduction}
\label{sec:introduction}

Refining the interpretation of images is a fundamental aim in computer vision \cite{lin2014microsoft}. It encompasses tasks such as image classification \cite{lin2014microsoft}, object detection \cite{redmon2016you}, and human pose estimation \cite{jin2020whole}. In sports, image interpretation has significantly contributed to foul detection \cite{tanaka2023automatic}, player identification \cite{suzuki2023runner}, and tracking \cite{scott2022soccertrack}. Recently, such computer vision techniques have been applied to analyze soccer posture effectively \cite{wear2022learning}.

In soccer posture analysis, a high-quality dataset and the analysis method hold profound effects on the analysis result. Concerning the pose dataset, for soccer pose and other human movements, the pose sequence is often captured in 2D given the use of a standard RGB camera \cite{mehraban2024motionagformer}. Currently, publicly available 2D pose datasets include MPII \cite{andriluka20142d}, OCHuman \cite{zhang2019pose2seg}, and COCO-WholeBody \cite{jin2020whole}. For soccer and sports image datasets, options include LSP \cite{johnson2010clustered}, Sports-102 \cite{bera2023fine}, LearningFromThePros \cite{wear2022learning}, and Sport Image \cite{wang2011learning}. These datasets have been effectively utilized for their designated tasks. Regarding the analysis method, existing studies have analyzed the goalkeeper \cite{wear2022learning,pinheiro2022body} and shooter \cite{nakamura2018differences,hong2011analysis} postures as they directly affect the match result. With a simplified scenario and linear model, studies have provided valuable insights into posture dynamics.

However, concerning the dataset, the large 2D pose datasets often lack a sufficient number of soccer images, and the soccer-specific datasets provide single images of postures instead of sequences, failing to represent entire movements. Regarding the analysis method, the simplified scenarios may not fully capture the complexities of real-world situations, limiting the applicability of results, especially at the highest competitive levels. Additionally, the utilization of linear models may overlook more intricate relationships in the data.

\begin{figure}[t]
  \centering
   \includegraphics[width=1\linewidth]{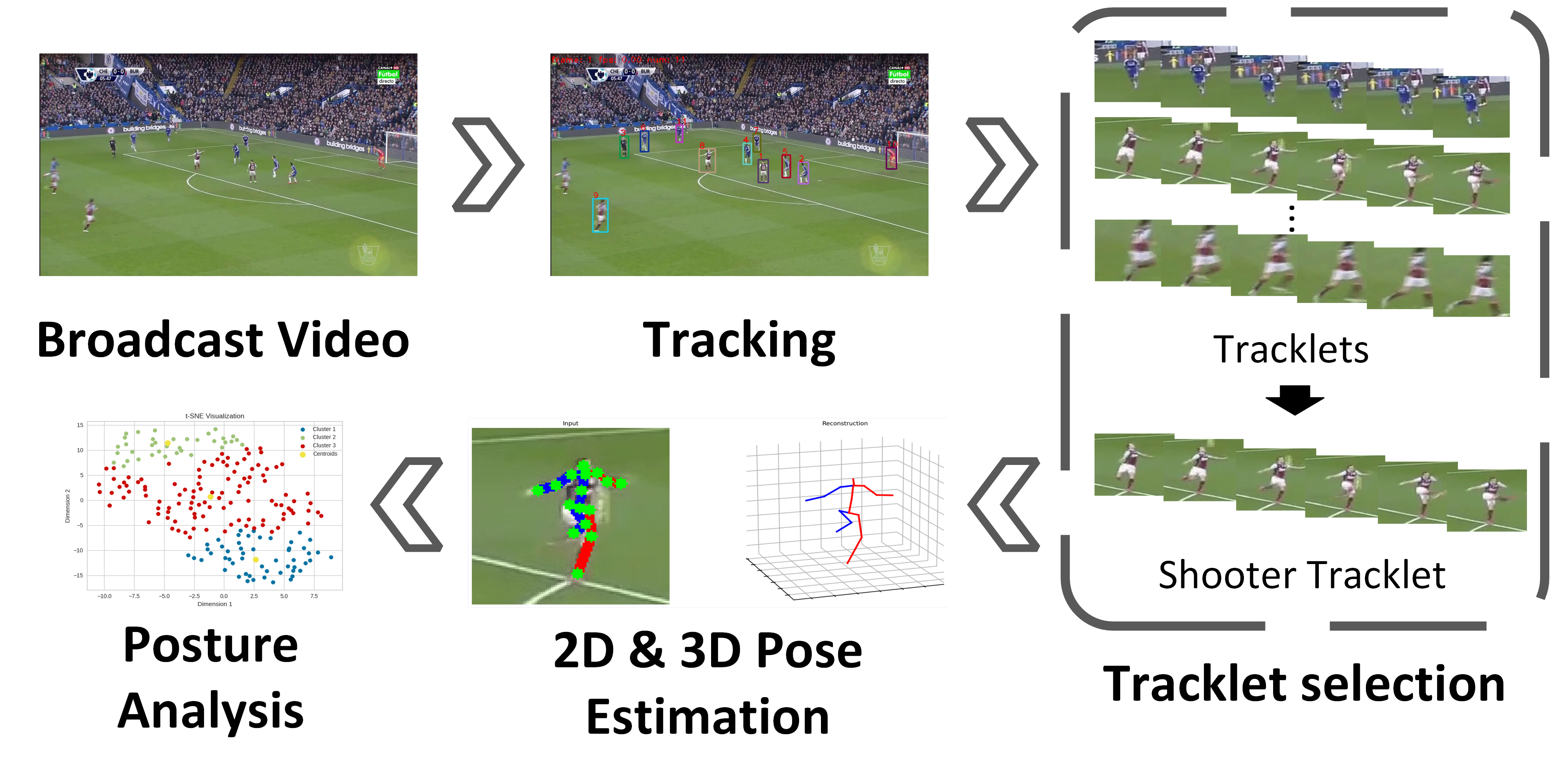}

   \caption{Overview of AutoSoccerPose.}
   \label{fig:overview_of_autosoccerpose}
\end{figure}

Therefore, in this study, we propose the 3D Shot Posture (3DSP) dataset, consisting of annotated 2D pose sequences of shooting instances from professional soccer matches. We introduce the 3DSP-GRAE (Graph Recurrent AutoEncoder) model, a non-linear model designed to encode pose sequences into vectors, and AutoSoccerPose, a semi-automated pipeline for extracting 2D and 3D pose sequences directly from professional soccer broadcast videos for analysis, as illustrated in \cref{fig:overview_of_autosoccerpose}. The contributions of this paper are as follows:

\begin{itemize}
\item 3DSP (3D Shot Posture) dataset, which comprises 2D pose annotations, 3D pose estimations, and tracklet information extracted from professional broadcast videos.
\item 3DSP-GRAE model, a novel nonlinear spatiotemporal model designed to encode pose sequences into vector representations effectively.
\item AutoSoccerPose, a pipeline tailored for semi-automated 3D posture analysis in soccer, specifically focusing on player shot movements.
\item Extensive evaluations at each stage of AutoSoccerPose, accompanied by an analysis of shooting postures utilizing the 3DSP dataset.
\end{itemize}

\section{Related work}
\label{sec:related_work}
\subsection{2D pose image datasets}
For images consisting of humans, 2D human pose estimation has emerged as a well-established research topic. This task aims to accurately define human body part segments or joint keypoints in the image. Commonly used benchmarks for this task include OCHuman \cite{zhang2019pose2seg}, COCO-WholeBody \cite{jin2020whole}, both derived from COCO \cite{lin2014microsoft}, and MPII \cite{andriluka20142d} dataset, covering various scenarios such as daily activities, human interactions, and sports events.

In the realm of sports, image frequently revolves around the posture of athletes. While traditional computer vision tasks such as image classification \cite{bera2023fine}, 2D/3D pose estimation \cite{johnson2010clustered,ingwersen2023sportspose}, and its extension, human parsing \cite{wang2011learning}, can be undertaken, the intricate anatomy involved, coupled with challenging image conditions like motion blur \cite{johnson2010clustered,bera2023fine}, amplify the difficulty of these tasks. In addition, advancements in computer vision techniques and soccer analytics, particularly those concerning player action value \cite{decroos2020vaep,yeung2023strategic,yeung2023transformer}, have opened avenues for leveraging soccer player posture in computer vision-based player evaluation \cite{wear2022learning}.
 
Nonetheless, while multiple datasets consist of pose images of professional soccer athletes, there are a few limitations when applying them for posture analysis:
\begin{enumerate}
    \item \textbf{General-purpose dataset}: Despite the potential abundance of images in conventional datasets, the proportion of professional soccer images is low due to the multitude of classes they encompass. Furthermore, these images do not specifically cater to distinct soccer actions, posing challenges for comparison.
    \item \textbf{Partial information}: Often, a complete soccer action, such as shooting, is captured through multiple sequential images (poses) \cite{hong2011analysis,nakamura2018differences}. However, current datasets only offer a single pose image for each discrete action. Moreover, 2D pose data relies on computer vision-based model estimation, introducing errors given the unique characteristics of sports images.
\end{enumerate}

To mitigate these issues, we have collected the 3DSP dataset (see \cref{sec:3dsp_dataset}), currently the largest repository of professional soccer human pose data with 2D posture annotations. This dataset focuses on the shooting action, providing sequential images of the shooting movement. In \cref{tab:overview_of_human_posture_in_sports_image_dataset}, we summarize the existing datasets containing professional soccer pose images along with the 3DPS dataset.


\begin{table*}[t!]
\centering
\begin{tabular}{l|cccccc}
\toprule
Dataset & \begin{tabular}[c]{@{}c@{}}2D Pose \\ Annotation\end{tabular} & \begin{tabular}[c]{@{}c@{}}Sequential \\ Images\end{tabular} & \begin{tabular}[c]{@{}c@{}}Images in \\ Soccer\end{tabular} & \begin{tabular}[c]{@{}c@{}}Number of \\ Images\end{tabular} & \begin{tabular}[c]{@{}c@{}}Soccer \\ Action\end{tabular} & Purpose \\  \midrule
LSP \cite{johnson2010clustered} & \cmark & \xmark & \naline & 2k & \naline & 2D Pose Estimation         \\
Sport Image \cite{wang2011learning} & \cmark & \xmark & \naline &1.3k & \naline & Human Parsing \\ 
MPII \cite{andriluka20142d}&\cmark & \xmark& 0.2k & 41k& \naline & 2D Pose Estimation\\
OCHuman \cite{zhang2019pose2seg}& \cmark & \xmark& \naline& 4.7K& \naline&2D pose estimation \\
COCO-WholeBody \cite{jin2020whole}& \cmark & \xmark& \naline& 200K& \naline&2D pose estimation\\
LearningFromThePros \cite{wear2022learning} & \xmark & \xmark & 1k &1k & GoalKeeping & Posture Analysis\\ 
Sports-102 \cite{bera2023fine} & \xmark & \xmark&0.2k&14k&\naline& Image Classification\\ 
\midrule
3DSP (Ours) & \cmark & \cmark & 4k & 4k & Shooting & Posture Analysis         \\ 
        \bottomrule
\end{tabular}
\caption{Overview of human pose in sports (professional soccer included) dataset. The datasets are arranged in order of their publication years. Due to the challenges associated with collecting 3D pose annotations during professional football matches, there currently exists no publicly available dataset providing 3D pose annotations for professional football.}
\label{tab:overview_of_human_posture_in_sports_image_dataset}
\end{table*}

\subsection{Posture analysis in soccer}
\label{ssec:posture_analysis_in_soccer}

Posture analysis holds profound importance in soccer, given its direct impact on player performance and strategic optimization. Previous studies have predominantly focused on shooting and goalkeeping postures, recognizing their pivotal roles in influencing match outcomes.

In shooting analysis, researchers have delved into unraveling the intricate differences between various kicking techniques. For instance, \cite{nakamura2018differences} examined disparities between non-rotational shots and instep kicks, while \cite{hong2011analysis} investigated distinctions among straight shots, knuckling shots, and curve shots. Both studies utilized advanced camera setups and reflective markers to capture 2D poses of college soccer players' shooting postures. The differentiation was based on 2D pose data, kinematic parameters, and principal component analysis techniques.

On the other hand, studies on goalkeeping posture, such as \cite{wear2022learning}, have focused on analyzing goalkeeper saving techniques in penalty and 1-vs-1 situations. Utilizing techniques like estimated 3D pose extracted with PoseHG3D \cite{zhou2017towards}, engineered features, and K-means clustering, researchers examined the various postures used by goalkeepers. Similarly, \cite{pinheiro2022body,pinheiro2021design} examined the strategies employed by goalkeepers in penalty kicks. They employed logistic regression to infer the relationship between orientation and strategies, where the orientation was engineered from the 2D pose of goalkeepers and penalty takers, utilizing OpenPose \cite{cao2017realtime} for pose estimation.

While previous studies provide valuable insight into the posture of soccer players, several drawbacks could limit the development of posture analysis:

\begin{enumerate}
\item \textbf{Simplified scenario}: Previous datasets are mainly collected under simplified scenarios, such as competitive settings with few players or controlled experimental environments. However, soccer actions often occur in occluded and crowded situations captured in broadcast videos.
\item \textbf{Labor-intensive}: Collecting data for posture analysis involves labor-intensive processes including image preprocessing, pose estimation or annotation, and quality control. Even for the datasets used in this study, the collection process had taken months to complete.
\item \textbf{Linear model}: Previous methods like principal component analysis and logistic regression offer valuable insights but may fail to capture non-linear spatiotemporal relationships present in posture data.
\end{enumerate}

Given these limitations, this study proposes the AutoSoccerPose for semi-automated 3D posture estimation from professional soccer broadcast video (see \cref{sec:proposed_method}) and a non-linear graph-based sequential model 3DSP-GRAE for posture analysis (see \cref{ssec:posture_analysis}).

\section{Proposed method}
\label{sec:proposed_method}
AutoSoccerPose\footnote{\label{autosoccerpose}For detailed specifications of the AutoSoccerPose, please refer to \url{https://github.com/calvinyeungck/3D-Shot-Posture-Dataset}.} aims to extract 3D poses from broadcast videos and analyze the extracted posture. Each stage of AutoSoccerPose corresponds to an existing computer vision task in soccer. \cref{fig:overview_of_autosoccerpose} provides an overview of the stages involved in AutoSoccerPose, and each stage is elaborated upon in the subsequent sections.

\subsection{Broadcast video}
\label{ssec:broadcast_video}
Similar to all posture analyses, Clips that capture the player's posture are fundamental to AutoSoccerPose. For broadcast video, there exist publicly available professional matches videos that could be collected, as in \cite{pinheiro2022body}, or existing publicly available datasets like SoccerNet \cite{deliege2021soccernet,giancola2018soccernet} that provide a large number of videos (nearly 800 hours of professional soccer matches). Nonetheless, while the posture of professional players in broadcast videos is the most informative and commonly available data, other less complex video footage like those in previous studies (see \cref{ssec:posture_analysis_in_soccer}) could be utilized in AutoSoccerPose.

Besides, accurate timestamp annotation of the action, shooting in this case, is critical to cutting clips from the broadcast video and the performance of AutoSoccerPose (see \cref{ssec:trackelt_selection_required_information}). The timestamp of important events like shooting could easily be retrieved in the post-match report, manually annotated, or publicly available datasets \cite{deliege2021soccernet,giancola2018soccernet,pappalardo2019public} as it was a fundamental type of soccer data annotation. Furthermore, accurate timestamp annotation has been a growing soccer computer vision task in recent times, in SoccerNet Challenge 2023 \cite{cioppa2023soccernet} to facilitate broadcast video understanding, the task action spotting (locate when and what type of event happened in the broadcast video) has been proposed. We reserve the integration of action spotting for future research.

To be more specific, the AutoSoccerPose was currently developed with the videos and timestamp annotation in SoccerNet \cite{deliege2021soccernet,giancola2018soccernet}, with frames resolution of $1280 \times 720$ and 25 fps. The retrieved clips used for AutoSoccerPose cover 0.5 seconds before and after the annotated timestamps. However via observation, the first 20 frames cover the entire shot movement for most of the shots, therefore only the first 20 were utilized for each shot.
\subsection{Tracking}
\label{ssec:tracking}
Tracking in AutoSoccerPose aims to identify soccer players in video clips and create tracklets: pixel coordinates of each respective player within each frame of the clips. Conventionally, two main tracking methods are utilized: tracking-by-detection \cite{Wojke2017simple,cao2023observation,du2023strongsort}, where objects are initially identified with bounding boxes in each frame and then associated (tracked) based on various conditions; and end-to-end tracking \cite{zhang2023motrv2,meinhardt2022trackformer,xu2020train}, where tracklets (bounding boxes of an object) are generated directly from the model.

State-of-the-art soccer player tracking models, achieving the best performance in the SoccerNet Challenge 2023, predominantly follow the tracking-by-detection approach. However, these tracking models are often not open-sourced. Here, we adopt the tracking by detection method using widely used models: fintuned YOLOv8\footnote{\label{yolov8}YOLOv8: \url{https://docs.ultralytics.com/}.} as the detector and BoTSort \cite{aharon2022bot} as the tracking algorithm. For performance evaluation and discussion of YOLOv8-BoTSort, refer to \cref{ssec:tracking_result_on_soccernet}.

\subsection{Tracklet selection}
\label{ssec:tracklet_selection}
Tracklet selection aims to pinpoint the tracklet associated with the shooter. This task can be viewed as a facet of image comprehension, where the objective is to classify whether a sequence of images pertains to the shooter. Analogous image comprehension tasks in soccer encompass sport type classification \cite{bera2023fine}, foul detection \cite{held2023vars,fang2024foul}, and action spotting \cite{cioppa2023soccernet}. Nonetheless, many existing studies propose intricate deep-learning architectures that necessitate substantial and diverse datasets for training or finetuning.

To avoid the need for an extensive dataset, we opt for a simpler approach employing a Convolutional Neural Network (CNN) model\textsuperscript{\ref{autosoccerpose}}. Our model processes the sequence of images through three standard convolutional blocks, each comprising a convolutional layer, ReLU activation function, max pooling layer, batch normalization, and dropout layer. Subsequently, the feature maps of the images within the sequence are concatenated. Finally, the concatenated feature maps are passed through a three-layer Multi-Layer Perceptron (MLP) with ReLU activation to predict whether the tracklet contains the shooter.

For each tracklet, we extract a sequence of images spanning frames 10 to 15. From each frame, we crop a tracklet image of size 96 $\times$ 96 pixels, with the center of the bounding box of the detected player (as described in \cref{ssec:tracking}) serving as the center of the tracklet image. The shooter tracklet for each clip is determined by selecting the tracklet with the highest estimated probability from the CNN model. Definitions of clips and frames can be found in \cref{ssec:broadcast_video}, while details regarding the selection of necessary frames are outlined in \cref{ssec:trackelt_selection_required_information}.

\subsection{2D \& 3D pose estimation}
\label{ssec:2d_and_3d_pose_estimation}
The objective of 2D and 3D pose estimation is to derive posture details from video clips, with a specific emphasis on capturing the movements of the shooter. 3D pose estimation methods can be categorized into direct 3D pose estimation \cite{chun2023learnable,iskakov2019learnable,reddy2021tessetrack} and 2D-3D lifting \cite{mehraban2024motionagformer,yang2023effective,sun2019deep}, which involves converting 2D pose coordinates into 3D space. The MPI-INF-3DHP dataset \cite{mehta2017monocular} is a widely used benchmark for in-the-wild 3D human pose estimation, resembling scenarios found in soccer broadcasts. Among the best-performing models, those employing the 2D-3D lifting approach, such as MotionAGFormer \cite{mehraban2024motionagformer}, are prevalent and thus adopted for 3D pose estimation in AutoSoccerPose.

For 2D pose estimation, methodologies are typically classified into bottom-up approaches \cite{cao2017realtime,cheng2020higherhrnet,geng2021bottom}, which detect points in the image and then associate keypoints to form 2D poses, and top-down approaches \cite{jiang2023rtmpose,sun2019deep,yang2023effective}, which first detect individuals using object detection models and then perform pose estimation within cropped bounding boxes. Top-down approaches, often more efficient, are favored, with RTMPose \cite{jiang2023rtmpose} selected for AutoSoccerPose due to its superior performance on datasets like COCO-whole body \cite{jin2020whole} and our dataset, 3DSP.

In implementation, for shooter tracklets, we extract sequences of 20 images, each sized 100 $\times$ 100 pixels, covering the detected players' bounding boxes and entire bodies, akin to \cref{ssec:tracklet_selection}. However, cropping images larger than the bounding box might include other players besides the shooter. To mitigate this, as the shooter is always centered in the cropped image, we select the shooter's 2D pose as the one with the center of body (torso) joint closest to the image center. The 2D pose from RTMPose \cite{jiang2023rtmpose} is then lifted to 3D using MotionAGFormer \cite{mehraban2024motionagformer}. The performance of both 2D and 3D pose estimation is further discussed in \cref{ssec:2d_pose_esimation_performance_on_3DSP}.

\subsection{Posture analysis}
\label{ssec:posture_analysis}

\begin{figure*}[t]
  \centering
   \includegraphics[width=0.8\linewidth]{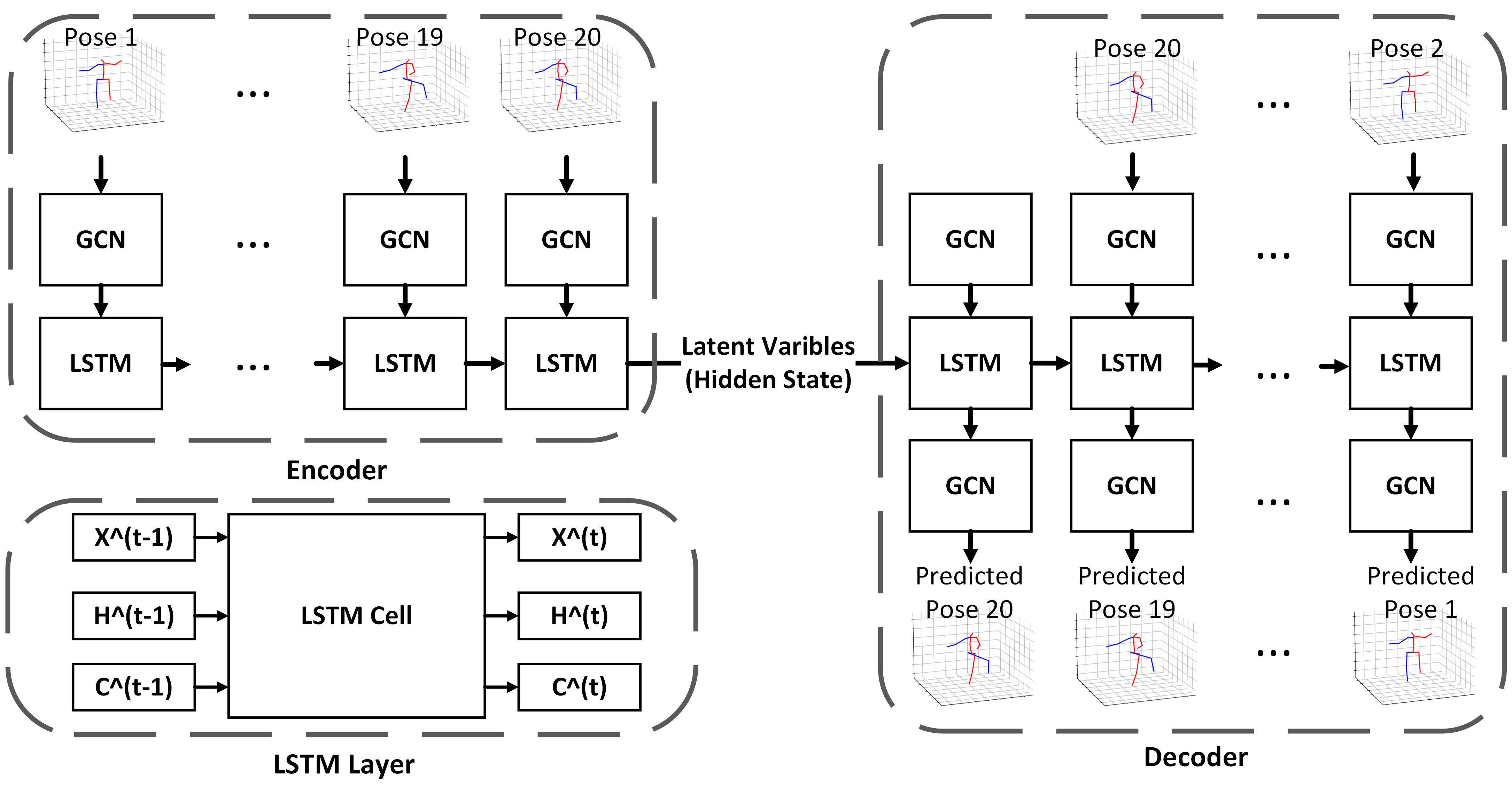}
   \caption{Overview of the 3DSP-GRAE model. The variables $X$, $H$, $C$, and $t$ in the LSTM layer denote the input features, hidden state, cell state, and time step, respectively. }
   \label{fig:g_lstm_ae_model}
\end{figure*}

Posture analysis seeks to distinguish various shooting styles among soccer players. Given that shot movements are typically unlabeled, unsupervised learning represents the most suitable approach for this task. Furthermore, as outlined in \cref{ssec:posture_analysis_in_soccer}, prior research \cite{hong2011analysis,pinheiro2022body} predominantly employs linear models such as Principal Component Analysis (PCA) and logistic regression to infer 3D posture, potentially limiting their ability to capture non-linear spatiotemporal relationships. Hence, to conduct unsupervised learning on posture data with a model capable of capturing the non-linear spatiotemporal properties, we propose the 3DSP-GRAE (Graph Recurrent AutoEncoder) model\textsuperscript{\ref{autosoccerpose}} to encode the latent (representation) vector, illustrated in \cref{fig:g_lstm_ae_model}. This model draws inspiration from the Graph Convolutional Network (GCN) \cite{kipf2016semi} and LSTM AutoEncoder (LSTM-AE) \cite{srivastava2015unsupervised}. While several graph and sequential-based models exist \cite{mehraban2024motionagformer,wang2020motion,ci2019optimizing}, none are explicitly designed for 3D pose encoding. Thus, leveraging the relatively simple structure of 3DSP-GRAE offers the advantage of requiring less data for training compared to fine-tuning existing models.

In general, the 3DSP-GRAE model utilizes an encoder-decoder architecture. For the encoder, 3D poses pass through a GCN to extract spatial relationships. Subsequently, the processed 3D poses are sequentially fed into an LSTM to extract nonlinear and temporal relationships. The output of the LSTM at the last layer and the last hidden state serve as the latent variables (vector representation) of the respective shot movement. In the decoder, another LSTM takes the latent variables and the processed 3D poses with GCN as input, but in reverse order. The LSTM output is then converted back to coordinates using another GCN. Consequently, the decoder output comprises the predicted 3D poses. Lastly, by employing the mean square error between the predicted 3D pose and the original input, the 3DSP-GRAE model can be trained.

In detail, each pose comprises 17 joints, as outlined in \cite{mehraban2024motionagformer}, with each joint represented by XYZ coordinates. The GCN consists of 2 simplified graph convolutional layers with ReLU activation, and the adjacency matrix represents connections between the joints with self-connections. Compared to the original graph convolutional layer \cite{kipf2016semi}, the parameter for the adjacency matrix is ignored, considering it to be the identity matrix. Furthermore, the encoder LSTM takes the pose input from pose 1 to 20 in temporal order. Meanwhile, in the decoder LSTM, the first input is zero-padded, followed by poses 20 to 2. 




\section{3DSP dataset}
\label{sec:3dsp_dataset}
In this section, we introduce the 3D Shot Posture (3DSP) dataset\footnote{\label{3dsp_dataset}3DSP dataset available at \url{https://github.com/calvinyeungck/3D-Shot-Posture-Dataset}.}, a crucial resource for developing AutoSoccerPose, serving as a benchmark for 2D pose estimation in professional soccer match broadcast videos. Furthermore, as evidenced by the comparative study outlined in \cref{tab:overview_of_human_posture_in_sports_image_dataset}, 3DSP comprises sequential images capturing soccer movements previously unavailable, with a specific focus on distinct soccer actions such as shooting. 3DSP also represents the largest collection of soccer images annotated with 2D pose information.
Additionally, the 3DSP dataset facilitates advancements in soccer posture analysis and offers direct applicability for future studies in this domain. Moreover, the 3DSP dataset could be extended easily with AutoSoccerPose.

\textbf{Data collection process}: 3DSP mirrors the process of AutoSoccerPose (with more details and code provided on GitHub \textsuperscript{\ref{autosoccerpose}}), albeit with manual refinement and annotation to ensure the highest quality of 2D poses. Initially, broadcast videos were sourced from SoccerNet \cite{deliege2021soccernet,giancola2018soccernet}, and utilizing SoccerNet annotations on actions, the videos were trimmed (0.5 seconds before and after the annotated timestamp, totaling 25 frames). Subsequently, tracklets for these clips were generated employing a fine-tuned YOLO v8\textsuperscript{\ref{yolov8}} in conjunction with BoT-Sort \cite{aharon2022bot}. Furthermore, the tracklet corresponding to the shooter was manually selected and refined, and a cropped image was produced using the bounding box. The 2D poses of the first 20 frames (determined empirically) were then manually annotated and lifted to 3D utilizing the MotionAGFormer \cite{mehraban2024motionagformer}.

To summarize the dataset structure, we collected data on shot movements from 22 unique matches in the 2015-2016 English Premier League. The training set comprises 20 cropped images (100 $\times$ 100 pixels) per shot movement, with a total of 200 shot movements, resulting in 4000 images. Annotations encompass tracklets, 2D poses, and estimated 3D poses. In the test set, there are also 20 cropped images per shot movement, with a total of 10 shot movements; however, only tracklet information is available. For both sets of data, references to SoccerNet \cite{deliege2021soccernet,giancola2018soccernet} are provided, along with the code utilized for extending 3DSP with SoccerNet \cite{deliege2021soccernet,giancola2018soccernet}.


\section{Experiments}
\label{sec:experiments}
In this section, we verify each step of the AutoSoccerPose (see \cref{sec:proposed_method}). More details is available at GitHub\textsuperscript{\ref{autosoccerpose}}.
\subsection{Tracking performance on SoccerNet \cite{deliege2021soccernet}}
\label{ssec:tracking_result_on_soccernet}

In this subsection, we assess the performance of AutoSoccerPose tracking using the test set provided in the SoccerNet Player Tracking Challenge 2023 \cite{cioppa2023soccernet}. This dataset comprises ground truth tracking annotations derived from broadcast videos in SoccerNet \cite{deliege2021soccernet,giancola2018soccernet}. We conduct evaluations for both player detection and tracking.

For player detection, we employ evaluation metrics including precision, recall, and average precision (AP) (see \cref{tab:soccer_player_detection_models_performance} caption). 
The evaluated models encompass state-of-the-art object detection models introduced subsequent to the SoccerNet Player Tracking Challenge 2023 \cite{cioppa2023soccernet}, including YOLOv8 \textsuperscript{\ref{yolov8}} and RT\_DETR \cite{lv2023detrs}, alongside their finetuned variants trained on the SoccerNet Player Tracking Challenge 2023 data \cite{cioppa2023soccernet}, utilizing the Python package Ultralytics. The outcomes are delineated in \cref{tab:soccer_player_detection_models_performance}. The result signified that YOLOv8 with finetuning yielded the most favorable player detection results. The relatively diminished recall rate before finetuning was attributed to player occlusion. However, with finetuning, YOLOv8 achieved a 97\% AP, in contrast to RT\_DETR \cite{lv2023detrs}, which exhibited an increase in AP but a decrease in precision.

\begin{table}[t]
\centering
\begin{tabular}{l|cc|c}
\toprule
Model               & Precision & Recall   & $\text{AP}_{0.5}$   \\
\midrule
YOLOv8\textsuperscript{\ref{yolov8}}              & 95.24\%   & 46.02\% & 80.33\% \\
RT\_DETR \cite{lv2023detrs}            & 92.01\%   & 50.83\% & 80.41\% \\
RT\_DETR \cite{lv2023detrs} (finetuned) & 85.00\%   & \textbf{94.30\%} & 85.10\% \\
YOLOv8\textsuperscript{\ref{yolov8}}  (finetuned)   & \textbf{95.70\%}  & 93.80\% & \textbf{97.30\%} \\
\bottomrule
\end{tabular}
\caption{Soccer player detection models performance. Ranked by AP, with the top-performing result highlighted in bold. Precision quantifies the proportion of correctly detected bounding boxes out of all detected ones, recall measures the proportion of correctly detected bounding boxes out of all ground truth boxes, and AP calculates the area under the precision-recall curve.}
\label{tab:soccer_player_detection_models_performance}
\end{table}

Concerning tracking evaluation, 
the compared methodologies encompass the baseline and the top-performing approach (Kalisteo \cite{maglo2023individual}) from the SoccerNet Player Tracking Challenge 2023 \cite{cioppa2023soccernet}, in addition to two prevalent tracking techniques integrated with the finetuned YOLOv8. The outcomes are depicted in \cref{tab:tracking_performance}. It was evident that the Kalisteo method \cite{maglo2023individual} continued to outperform popular methods. However, the absence of publicly available code for the Kalisteo method \cite{maglo2023individual} posed a limitation. Furthermore, the test set of the SoccerNet Player Tracking Challenge 2023 \cite{cioppa2023soccernet} comprised numerous clips featuring substantial player occlusion, whereas in most shooting scenarios, player occlusion was less pronounced. Hence, we ascertain that BoTSort suffices for AutoSoccerPose.

\begin{table}[t]
\centering
\begin{tabular}{l|cc|c}
\toprule
Method             & DetA    & AssA    & HOTA    \\
\midrule
SoccerNet Baseline \cite{cioppa2023soccernet} & 38.38\% & 49.81\% & 43.67\% \\
YOLOv8\textsuperscript{\ref{yolov8}}-ByteTrack \cite{zhang2022bytetrack}   & 63.75\% & 42.31\% & 51.85\% \\
YOLOv8\textsuperscript{\ref{yolov8}}-BoTSort \cite{aharon2022bot}    & 68.80\% & 53.40\% & 60.54\% \\
YOLOX \cite{ge2021yolox}-Kalisteo \cite{maglo2023individual}     & \textbf{73.64}\% & \textbf{73.76\%} & \textbf{73.65\%} \\
\bottomrule
\end{tabular}
\caption{Tracking performance on SoccerNet player tracking challenge 2023 \cite{cioppa2023soccernet} test set. The evluation metrics include DetA (Detection Accuracy), AssA (Association Accuracy), and HOTA (Higher Order Tracking Accuracy) \cite{luiten2021hota}. Ranked by HOTA, with the top-performing result highlighted in bold.}
\label{tab:tracking_performance}
\end{table}

\subsection{Tracklet selection required features}
\label{ssec:trackelt_selection_required_information}

\begin{table}[t]
\centering
\begin{tabular}{l|cc|cc}
\toprule
\begin{tabular}[c]{@{}c@{}}Required\\ Frames\end{tabular} & Precision        & Recall           & ACC              & \begin{tabular}[c]{@{}c@{}}CLIP\\ ACC \end{tabular}\       \\
\midrule
Frame 12-13       & 20.81\%          & \textbf{77.50\%} & 76.52\%          & 52.50\%          \\
Frame 11-14         & 38.33\%          & 57.50\%          & 90.02\%          & 65.00\%          \\
Frame 10-15         & \textbf{57.78\%} & 65.00\%          & \textbf{93.90\%} & \textbf{75.00\%} \\
Frame 9-16          & 41.82\%          & 57.50\%          & 90.94\%          & 65.00\%     \\
\bottomrule
\end{tabular}
\caption{Tracklet selection model performance utilizing different numbers of frames. Ranked by the number of required frames, with the top-performing result highlighted in bold.}
\label{tab:tracklet_selection-performance}
\end{table}

\begin{table*}[t]
\centering
\begin{tabular}{l|cccccccc|cc}
\toprule
Model  & Head    & Sho     & El      & Wri     & Body    & Hip     & Knee    & Ank     & PDJ     & AUC     \\
\midrule
HRNet \cite{sun2019deep}   & 94.56\% & 63.38\% & 21.14\% & 17.29\% & 96.35\% & 78.68\% & 47.46\% & 34.94\% & 56.08\% & 47.38\% \\
DWPose \cite{yang2023effective}  & 86.69\% & 85.80\% & 72.75\% & 58.88\% & 89.88\% & 71.41\% & 45.41\% & 32.98\% & 67.94\% & 55.80\% \\
RTMPose \cite{jiang2023rtmpose}& \textbf{96.81\%} & \textbf{95.08\%} & \textbf{86.22\%} & \textbf{76.01\%} & \textbf{96.98\%} & \textbf{95.62\%} & \textbf{88.30\%} & \textbf{79.00\%} & \textbf{89.51\%} & \textbf{73.56\%} \\
\bottomrule
\end{tabular}
\caption{2D pose estimation performance on 3DSP. Ranked by PDJ, with the top-performing result highlighted in bold. Sho, El, Wri, and Ank represent Shoulder, Elbow, Wrist, and Ankle respectively. Each column for body parts indicates the average PDJ for the specified body part, while the PDJ column denotes the mean PDJ across all 17 keypoints.}
\label{tab:2d_pose_performance}
\end{table*}

\begin{figure*}[t]
  \centering
   \includegraphics[width=0.8\linewidth]{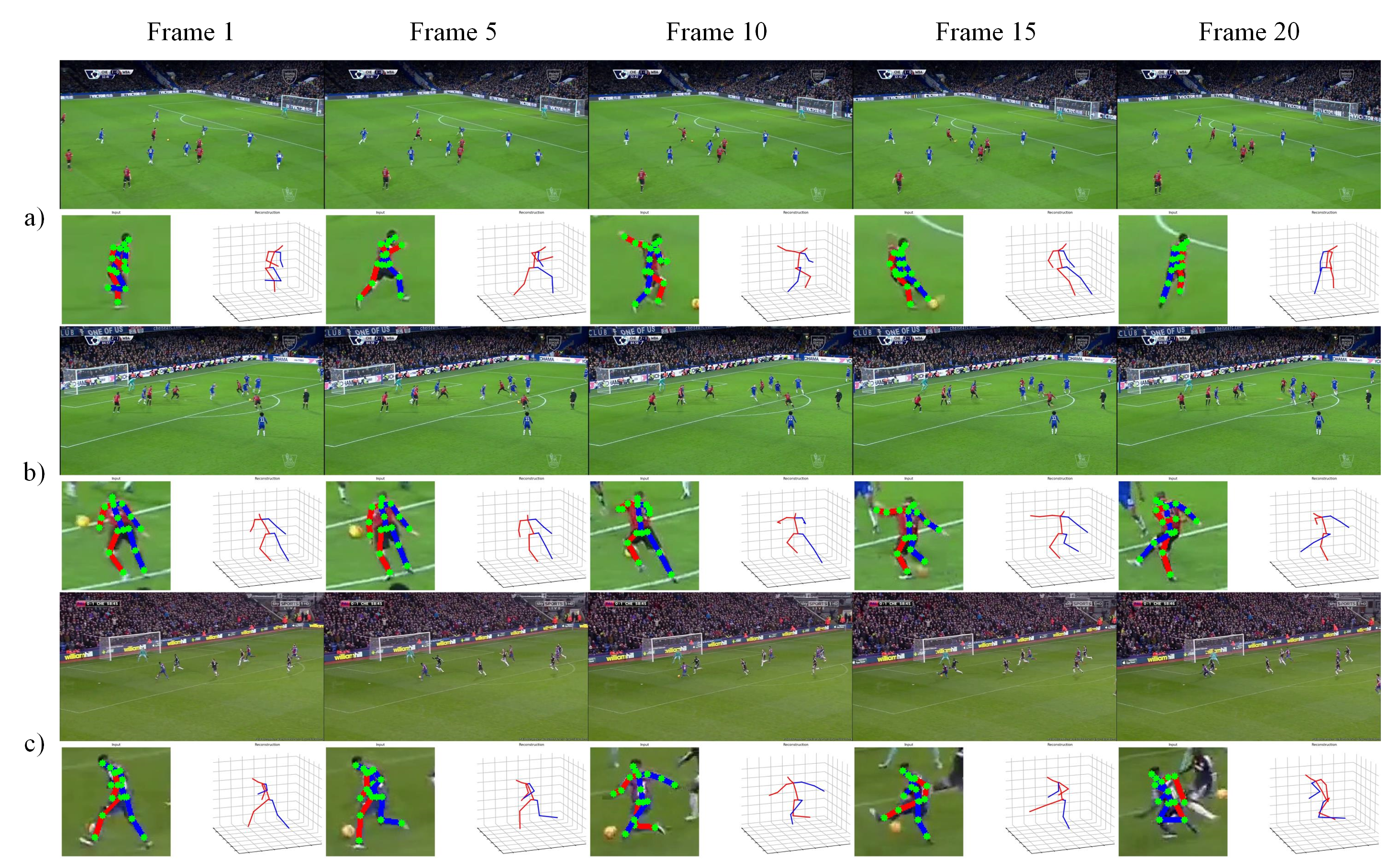}
   \caption{AutoSoccerPose qualitative results.  Each row depicts the AutoSoccerPose 2D and 3D pose estimation from a broadcast video, and the columns denote which frame of the video. The top photo represents the broadcast video frame for each cell, while the bottom-left and bottom-right images correspond to the 2D and 3D pose estimations, respectively.}
   \label{fig:qualitative_result}
\end{figure*}

After verifying the tracking, the next crucial step is tracklet selection, performed using a CNN model in AutoSoccerPose (see \cref{ssec:tracklet_selection}). Given the absence of existing models tailored for this specific task, we focus on validating essential features, namely the frames containing pivotal information. Our evaluation criteria encompass precision and recall metrics,
alongside accuracy (ACC) and clip accuracy (CLIP ACC).
The ACC metric assesses the CNN model's accuracy in classifying a tracklet as the shooter's tracklet. Additionally, CLIP ACC evaluates the accuracy of tracklet selection, specifically in identifying the shooter's tracklet from all tracklets within a clip (see \cref{ssec:tracklet_selection}).
200 clips are sourced from SoccerNet \cite{deliege2021soccernet,giancola2018soccernet}, complemented by tracklet information from the 3DSP train set (see \cref{sec:3dsp_dataset}).

The feature set is based on tracklet images spanning frames 12 and 13, bounding the timestep of the shot movement (see  \cref{sec:proposed_method}). We iteratively enrich the feature set by progressively incorporating the frames preceding and succeeding the aforementioned frames. For each feature set, the CNN model is trained using PyTorch with an 80/20 train-validation split. Grid searching is employed to optimize the number of convolutional layers, MLP layers, and the hidden size of the MLP layer. The results are summarized in Table \cref{tab:tracklet_selection-performance}. Notably, employing tracklet images from frames 10 to 15 yields the best performance. While the tracklet selection model demonstrates reasonable performance, there remains potential for further enhancement.

\subsection{2D pose estimation performance on 3DSP}
\label{ssec:2d_pose_esimation_performance_on_3DSP}

This subsection aims to evaluate the performance of 2D pose estimation models under zero-shot conditions. We utilized the image and 2D annotations from the 3DSP dataset train set (see \cref{sec:3dsp_dataset}). The dataset provides annotations for 17 keypoints representing various body joints. To assess the performance, we employed the Percent of Detected Joints (PDJ) evaluation metric \cite{toshev2014deeppose}. In this metric, a keypoint is considered detected if the normalized distance between the predicted and ground truth keypoints is under 0.5. Additionally, the area under the PDJ curve (AUC) was calculated to report the performance across different thresholds.

Given the dynamic nature of the shooting movements, especially significant rotation of the shoulders and hips during shots, the distance between the contralateral shoulder and hip keypoints may not accurately represent the torso's largest distance consistently. Therefore, we normalized the distance between detected keypoints and ground truth keypoints by the distance between the center of the shoulder and the center of the hip, as suggested in \cite{tompson2014joint}.

The evaluated models include DWPose \cite{yang2023effective} and RTMPose \cite{jiang2023rtmpose}, which are the top-performing models in the COCO-WholeBody dataset \cite{jin2020whole} for 2D pose estimation tasks. Additionally, we included HRNet \cite{sun2019deep}, which is the 2D pose estimation model integrated into MotionAGFormer \cite{mehraban2024motionagformer}. The results were summarized in \cref{tab:2d_pose_performance}, indicating that RTMPose \cite{jiang2023rtmpose} exhibited the best performance across all joint detections. The detection errors predominantly occurred in limb detection. Through observation, two primary reasons were identified: when the shooter faced backward, the left and right limbs could be misidentified, and when the shooter faced sideways, one side of the limbs was covered by the torso.

\subsection{3D pose estimation on non-annotated data}

Here, we validate the results obtained by AutoSoccerPose for estimating 3D poses from broadcast videos, instead of relying on 2D annotations from 3DSP. This approach directly showcases what AutoSoccerPose users can retrieve without annotated data. Qualitative results are depicted in \cref{fig:qualitative_result}, comprising the outcomes of three clips: a) a shot towards the goal on the right from outside the box, b) a shot towards the goal on the left from outside the box, and c) a shot towards the goal on the left from inside the box. While the majority of 2D and 3D pose estimations were satisfactory, two limitations of AutoSoccerPose were identified: motion blur, as in Clip (a) Frame 15, which impeded the accurate detection of limb keypoints, and occlusion between the shooter and other players, as in Clip (b) Frame 20, hindered the detection of limb positions.

\subsection{Soccer shots posture analysis}
\label{ssec:soccer_shots_posture_analysis}

\begin{figure}[t]
  \centering
   \includegraphics[width=0.8\linewidth]{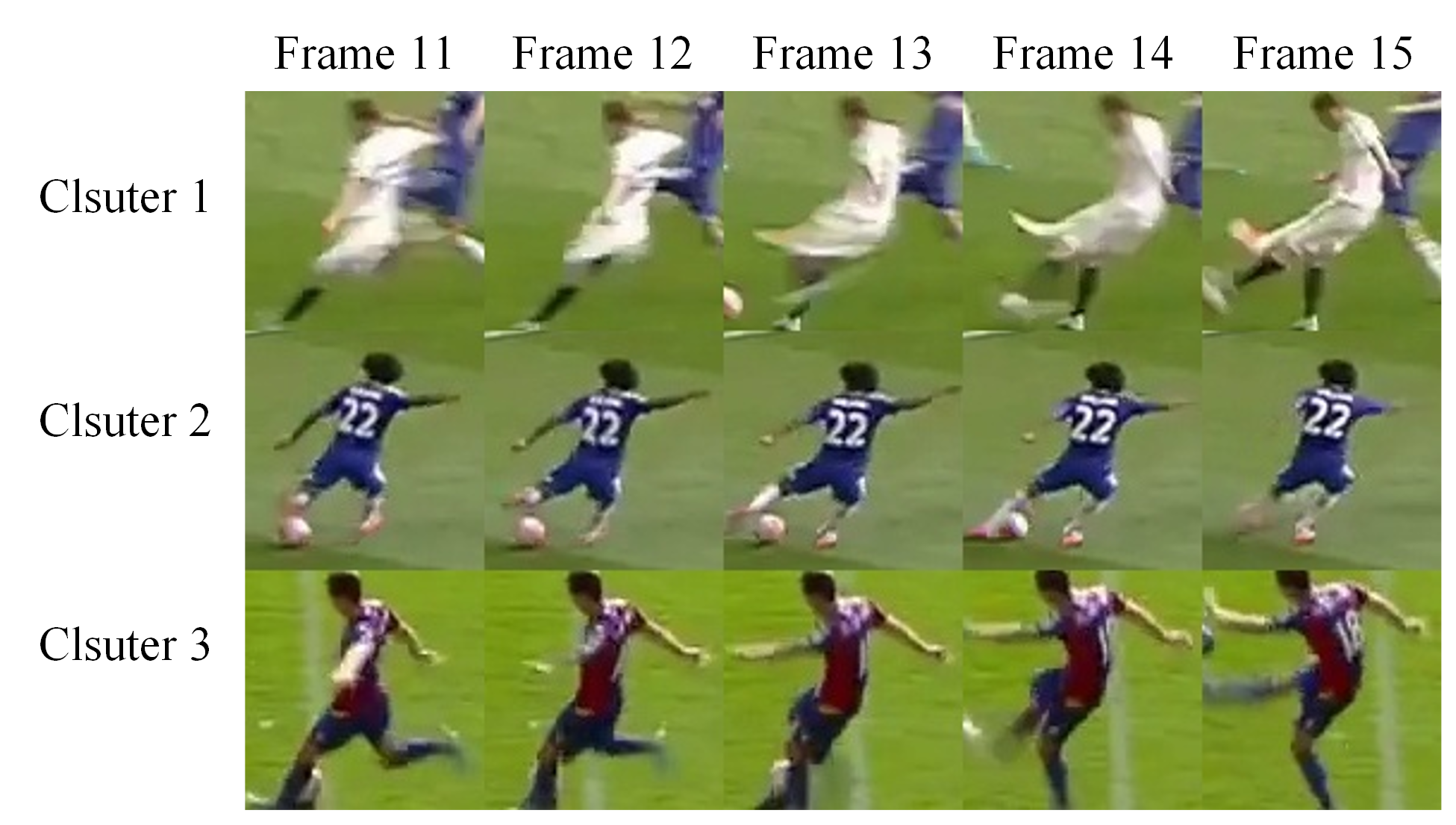}
   \caption{Shot movement closest to the cluster centroid. The row and column denote the cluster and frame, respectively.}
   \label{fig:shot_movement_closest}
\end{figure}

This subsection is dedicated to analyzing the 3D posture of professional soccer players. We leverage the 3D pose data available in the 3DSP dataset (see \cref{sec:3dsp_dataset}) to ensure high data fidelity for insightful analysis. Our goal is to identify various shooting techniques. To achieve this, we employ the 3DSP-GRAE model (see \cref{ssec:posture_analysis}) to extract representative vectors from sequential shot movements (comprising 3D postures from 20 frames). Subsequently, we apply K-means clustering \cite{arthur2007k} to categorize these vectors. The resulting clusters are visualized in a 2-dimensional space using t-distributed stochastic neighbor embedding (t-SNE) \cite{van2008visualizing} and illustrated in \cref{fig:cluster_fig}. Furthermore, we showcase the shot movements closest to each cluster centroid by displaying frames 11 to 15 in \cref{fig:shot_movement_closest}. Based on observation, Cluster 1 and Cluster 2 represent inside shots, whereas Cluster 3 denotes instep shots. The definitions are as follows:

\begin{figure}[t]
  \centering
   \includegraphics[width=0.7\linewidth]{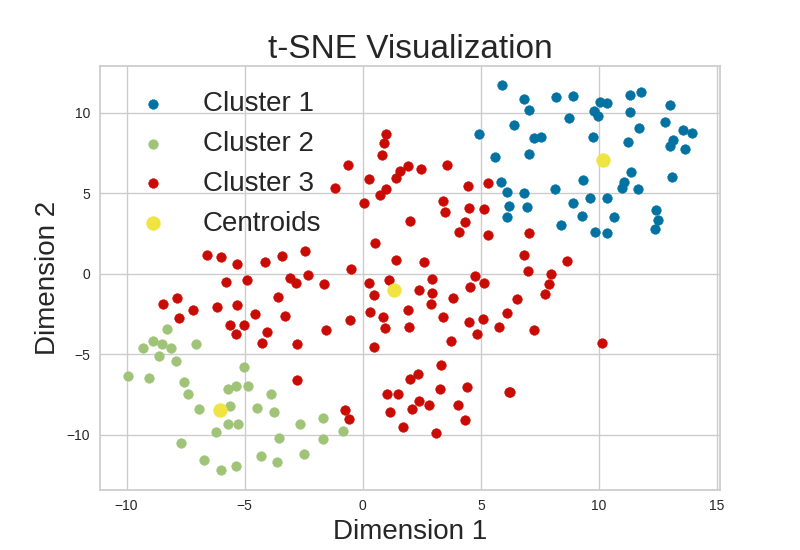}

   \caption{K-means clustering visualization.}
   \label{fig:cluster_fig}
\end{figure}

\begin{itemize}
\item \textbf{Inside shot}: Precise technique where the ball is struck with the inside of the foot, allowing for accurate placement into goal corners. Involves swinging the foot inward toward the opposite foot for finesse shots.
\item \textbf{Instep shot}: Powerful technique involving striking the ball with the area of the foot containing the laces. Commonly used for long-range or forceful shots. Players approach the ball straight on and use the laces to generate significant power.
\end{itemize}
Comparing Cluster 1 and 2 with statistics, Cluster 1 exhibited a 16\% greater shooting feet ankle average travel distance. Moreover, the maximum vertical coordinate in Cluster 1 surpassed that of Cluster 2 by 33\%, while the minimum knee angle was 17\% lower. Statistics revealed that shots in Cluster 1 generated a larger swing motion compared to those in Cluster 2, likely attributable to deeper knee bending and a more pronounced vertical trajectory. This enabled enhanced momentum transfer and force generation during the shot. The identified shooting style holds promise for further utilization in player evaluation for optimizing posture and style, as demonstrated in \cite{wear2022learning}. Meanwhile, when the aforementioned analysis was performed with linear model PCA and K-means \cite{arthur2007k}, the clustering result could only reflect shots to the left, center, and right, demonstrating the need for a non-linear spatiotemporal model (see GitHub\textsuperscript{\ref{autosoccerpose}}).

\section{Conclusion}
\label{sec:conclusion}
In this paper, we introduce AutoSoccerPose, a framework designed for estimating both 2D and 3D shooting postures of soccer players and conducting comprehensive posture analysis. Additionally, we present the 3DSP dataset, which stands as the largest annotated 2D pose dataset for professional soccer images. Through experimental validation, we showcase the performance of AutoSoccerPose across each stage of the framework, culminating in an assessment of its overall efficacy. Given that each component of AutoSoccerPose corresponds to a well-established task in computer vision, the seamless integration of future state-of-the-art models into AutoSoccerPose would readily enhance its capabilities. We envision that our research will inspire automated posture analysis in soccer and various other sports.
\section{Acknowledgement}
This work was financially supported by JSTSPRING GrantNumber JPMJSP2125 and JSPS KAKENHI 23H03282. 
{
    \small
    \bibliographystyle{ieeenat_fullname}
    \bibliography{main}
}


\end{document}